\documentclass[conference]{IEEEtran}
\usepackage{cite}
\usepackage{amsmath,amssymb}
\usepackage{graphicx}
\usepackage{url}
\usepackage{booktabs}
\usepackage{listings}
\usepackage{algorithm}
\usepackage{algpseudocode}

\usepackage[dvipsnames]{xcolor}

\title{Machine Learning-Based Graph Simplification \\for Symbolic Accelerators}

\author{
\IEEEauthorblockN{Tiffany Yu, Rye Stahle-Smith, Darssan Eswaramoorthi, Rasha Karakchi}
\IEEEauthorblockA{University of South Carolina\\
Email: \{tyu, ryes, darssan\}@email.sc.edu, karakchi@cec.sc.edu}
}

\begin{document}
\maketitle

\begin{abstract}
Graph-based accelerators have been widely adopted in symbolic data processing applications such as genomics, cybersecurity, and artificial intelligence. However, these systems often suffer from excessive memory usage and inefficiencies stemming from redundant graph structures. We present AutoSlim, a machine learning-based framework that leverages data-driven methods to prune automata graphs for hardware accelerators. Using features extracted from prior graph executions and a Random Forest classifier, AutoSlim identifies and removes low-impact nodes and edges. When applied to a Non-deterministic Finite Automata overlay architecture (NAPOLY+), AutoSlim reduces FPGA resource usage by up to 40\%, with corresponding improvements in throughput and power efficiency. The framework includes a verification step to ensure functional equivalence after pruning and suggests promising directions for both hardware optimization and security.

\end{abstract}


\section{Introduction}
Symbolic data—such as genomic sequences, text tokens, and behavioral logs—are prevalent across domains such as bioinformatics, cybersecurity, and natural language processing. These data types are often modeled and processed using finite automata, where patterns are encoded as graphs composed of states (nodes) and transitions (edges)~\cite{woods2018automata}. Finite state machines provide a powerful abstraction for pattern matching and symbolic analysis, forming the computational core of many practical applications.

While traditional CPU- and GPU-based implementations of automata processing have made strides in software optimization, their performance remains fundamentally limited by irregular memory access patterns and constrained parallelism. To address these challenges, the research community has developed domain-specific hardware accelerators that leverage FPGAs and ASICs for parallel, high-throughput automata execution~\cite{karakchi2019overlay, karakchi2016high, karakchi2017reconfigurable, karakchi2024transformer, karbowniczak2024scored, karbowniczak2025optimizing}.

Among these, NAPOLY+ stands out as a scalable overlay architecture for implementing Non-deterministic Finite Automata (NFA) on FPGAs. NAPOLY+ accepts automata described in the ANML format and maps their components directly to hardware: transitions are stored in on-chip BRAMs, and states are realized as point-to-point connected State Transition Elements (STEs)~\cite{karakchi2023napoly}. Its support for scoring-based pattern matching, in which the minimal-cost path determines the match relevance, makes it particularly effective for ranking-based applications. However, as the scale and complexity of symbolic datasets continue to grow, such systems face mounting scalability and efficiency constraints.

In particular, large automata graphs result in:
\begin{itemize}
    \item \textbf{Excessive memory usage}: each node and transition consumes FPGA resources, often leading to capacity overflow,
    \item \textbf{Routing congestion}: point-to-point interconnects become increasingly dense, complicating place-and-route phases,
    \item \textbf{Redundant computation}: unreachable states and duplicated structures waste logic without contributing to correctness.
\end{itemize}

To mitigate these bottlenecks, we propose AutoSlim, a machine learning-based framework for optimizing automata graphs prior to hardware deployment. AutoSlim operates during the graph preprocessing stage, with the aim of reducing the graph’s size and complexity while preserving the original automaton’s semantics and output behavior.

\begin{figure}[t]
\centering
\includegraphics[width=0.3\textwidth]{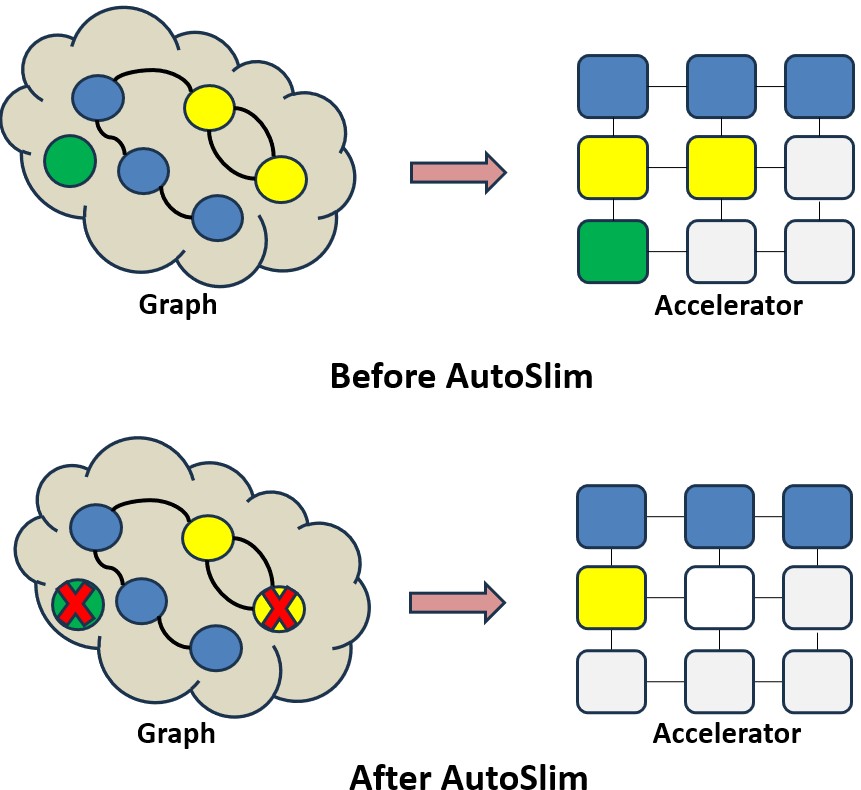}
\caption{Illustration of AutoSlim's graph optimization. \textit{Top:} Original automaton graph, where duplicated destinations (yellow nodes) and unreachable states (green) increase PE count and routing complexity. \textit{Bottom:} AutoSlim-pruned graph with merged redundant nodes and eliminated unreachable ones, enabling more efficient FPGA deployment.}
\label{fig:system}
\end{figure}

Figure~\ref{fig:system} illustrates AutoSlim’s core motivation. The top diagram (“Before AutoSlim”) shows an automaton with five nodes, two of which (in yellow) represent the same logical destination and thus unnecessarily consume multiple Processing Elements (PEs). Mapping this graph directly to hardware results in redundant resource allocation. AutoSlim identifies such semantically equivalent structures and merges them, reducing PE usage. Similarly, unreachable states (e.g., the green node) are pruned out since they cannot be activated during matching. These optimizations significantly reduce the hardware footprint without affecting correctness.

Empirical analysis of real-world workloads reveals that automata graphs often include substantial redundancy: unused transitions, unreachable nodes, and duplicated paths. Traditional graph reduction techniques—such as partitioning \cite{nourian2017demystifying}, state merging \cite{wadden2017automata}, and structural heuristics—often \cite{karakchi2023napoly, karakchi2019overlay} introduce trade-offs such as reduced interpretability or increased hardware complexity. Moreover, they tend to ignore the probabilistic nature of symbolic workloads, where certain nodes and transitions are statistically more significant than others.

AutoSlim addresses these challenges through a learning-based pruning pipeline. It extracts structural and behavioral features from annotated automata graphs—including edge frequency, transition cost, and node centrality—and uses a Random Forest classifier~\cite{breiman2001random} to predict the utility of each component. Edges and nodes deemed redundant are removed, and the resulting graph is validated to ensure that all critical accepting paths remain intact.

Unlike static rule-based methods, this data-driven approach generalizes across domains and workloads. Crucially, AutoSlim is designed to be fully compatible with the NAPOLY+ scoring-based matching model. Pruned graphs maintain the functional properties required for accurate symbolic processing and sequence ranking.

In summary, AutoSlim introduces a principled and hardware-aware methodology for automata graph optimization. By reducing graph size, simplifying routing, and eliminating unnecessary hardware utilization, it directly addresses key bottlenecks in FPGA-based symbolic accelerators. Furthermore, AutoSlim lays the groundwork for integrating security analysis—such as detecting adversarial or malicious graph components—thus enabling future research in robust and trustworthy automata processing systems.

\section{Background and Related Work}

\subsection{Automata Acceleration on Reconfigurable Platforms}

Finite automata are widely used in symbolic processing tasks such as pattern matching, sequence alignment, and intrusion detection \cite{karakchi2023napoly, karakchi2024transformer}. Traditional CPU- and GPU-based automata execution often suffers from memory inefficiencies and limited parallelism due to the irregular structure of automata graphs. To address this, several efforts have focused on mapping automata to spatial computing platforms, particularly FPGAs and custom architectures like Micron’s Automata Processor (AP).

\subsection{NAPOLY+: Score-Aware Automata Execution}

NAPOLY+ is a domain-specific overlay for FPGA-based acceleration of Non-deterministic Finite Automata (NFA)~\cite{karbowniczak2024scored, karbowniczak2025optimizing}. It represents automata as weighted graphs, where each edge carries a symbolic transition and a domain-specific cost. This design allows NAPOLY+ to compute the best match not only by detecting pattern acceptance, but also by identifying the minimum-cost path through the automaton. This score-aware approach is essential in applications such as ranked sequence alignment, approximate pattern matching, and weighted token matching in NLP pipelines. Unlike earlier spatial automata processors that rely on binary matching and fixed control flow, NAPOLY+ treats automata as dynamic, probabilistic graphs. Each state is instantiated as a State Transition Element (STE), and transitions are mapped to configurable routing and BRAM blocks. While this design offers high throughput and flexible scoring, it also introduces new challenges in resource consumption, routing congestion, and synthesis time—especially when handling large graphs with thousands of nodes and non-uniform edge weights.

\subsection{Limitations of Prior Work}

Previous work on automata acceleration has made substantial contributions in characterizing hardware platforms and developing toolchains \cite{rahimi2020grapefruit, wang2023hap, kobori2001cellular, subramaniyan2017parallel, zhang2022fra}, but does not fully address the scalability and score-aware complexity seen in practical applications.

Nourian \textit{et al.}~\cite{nourian2017demystifying} provide a comprehensive evaluation of automata execution across GPUs, FPGAs, and the Micron AP. Their study demonstrates that FPGAs offer favorable energy efficiency and throughput. However, the experiments are constrained to unit-cost automata and small-scale benchmarks from the ANMLZoo suite \cite{wadden2016anmlzoo}. The authors do not explore optimization of the automaton itself; their pipeline assumes a fixed input graph and focuses solely on execution backends.

Similarly, Wadden \textit{et al.}~\cite{wadden2017automata} introduce Automata-to-Routing, a spatial toolchain for mapping automata to hardware fabrics. Their framework focuses on optimizing routing strategies and architectural design parameters. However, similar to the approach of Nourian \textit{et al.}~\cite{nourian2017demystifying}, their tool assumes a static automaton and does not consider the impact of transition frequency, scoring, or semantic redundancy within the graph.

In contrast, our proposed AutoSlim framework targets the front end of the design flow. Rather than optimizing the hardware mapping of a fixed graph, AutoSlim performs graph simplification through machine learning-guided pruning. It analyzes transition cost, frequency, and structural connectivity to identify low-impact nodes and edges that can be safely removed—reducing resource usage while maintaining scoring accuracy. This approach is both data-aware and hardware-aware, aligning with the semantic model of NAPOLY+.

\subsection{AutoSlim: A Scalable and Semantic-Aware Approach}

AutoSlim differs fundamentally from prior work by targeting large automata graphs encountered in practice that incorporate scoring semantics. In symbolic workloads such as DNA alignment or weighted rule matching, the ability to find the least-cost match path is critical. AutoSlim respects these semantics by preserving essential scoring paths while pruning unreachable or redundant subgraphs.

Moreover, AutoSlim generalizes beyond small synthetic benchmarks, addressing the scalability gap left by previous studies. While Nourian and Wadden validate their techniques on ANMLZoo-scale automata, AutoSlim is designed for graphs with tens of thousands of nodes, incorporating statistical variation in edge importance. By integrating AutoSlim with NAPOLY+, we enable a complete pipeline—from semantic-aware graph reduction to efficient spatial hardware execution—tailored to symbolic workloads that demand both correctness and scalability.

\section{AutoSlim Tool and Methodology}

AutoSlim introduces a modular and scalable software framework designed to generate, annotate, and intelligently prune automata graphs for acceleration on FPGA platforms. The toolchain supports the entire optimization workflow, from dataset generation to hardware-ready graph deployment. This section describes the AutoSlim methodology in two main stages: symbolic graph dataset generation and learning-based pruning.

\subsection{Symbolic Graph Dataset Generation}

To support machine learning-based graph pruning, we first developed a symbolic dataset generator compatible with the NAPOLY+ automata overlay format~\cite{karakchi2023napoly}. Our tool builds on the ANMLZoo format—widely used in automata processing research—but extends it by embedding scoring metrics into each transition. Each graph is composed of state transition elements (STEs), and every transition is annotated with a numerical score representing its importance, frequency, or computational cost.

The generator supports user-defined graph sizes and transition densities, enabling the creation of both synthetic and biologically inspired graph topologies. Internally, the graphs are encoded in XML format for NAPOLY+ compatibility. These XML files are then parsed and converted to CSV representations containing structured fields such as node IDs, transition degrees, and cumulative edge scores. This transformation prepares the data for feature extraction and subsequent use in training machine learning models.

\begin{algorithm}[t]
\caption{ML-based Transition Pruning Pipeline}
\begin{algorithmic}[1]
\State \textbf{Input:} Folder of XML/CSV files with transitions, threshold $\theta$
\State \textbf{Output:} Pruned transition files and evaluation reports

\Procedure{PruneTransitions}{inputFolder, outputFolder, $\theta$}
    \ForAll{XML files in \texttt{inputFolder}}
        \State Parse each transition $(from, to, score)$
        \State Update node degree and total node score
        \State Export data to CSV format
    \EndFor
    \State Create labeled dataset $D = \{(x_i, y_i)\}$, where $y_i = \mathbb{1}[x_i > \theta]$
    \State Train Random Forest classifier on $D$
    \ForAll{CSV files}
        \State Classify each transition $(x_i)$ using the trained model
        \State Keep transitions with $\hat{y}_i = 1$
        \State Save pruned transitions to output CSV
        \State Generate report with state/transition count and accuracy
    \EndFor
    \State \Return Pruned files and evaluation reports
\EndProcedure
\end{algorithmic}
\end{algorithm}

\subsection{Transition Pruning via Supervised Learning}

The core of AutoSlim lies in its learning-based pruning pipeline, which reduces graph complexity by identifying and removing redundant or low-impact transitions. Using the CSV files generated from the symbolic graphs, AutoSlim constructs feature vectors for each transition. The primary feature used in the current implementation is the edge score, although additional graph features such as node degree or path centrality can be incorporated in future versions.

We employ a supervised Random Forest classifier~\cite{breiman2001random} to determine which transitions can be safely pruned without significantly affecting the graph’s matching accuracy. Labels for training are automatically derived from score thresholds, allowing flexible and dataset-specific cutoff tuning. The model is trained on a subset of transitions and evaluated on the remainder using standard cross-validation techniques.

Each graph is then passed through the trained model to filter out transitions predicted to be non-essential. The pruned graphs are saved in the same format as the originals, facilitating direct comparison and deployment.

\subsection{Reporting and Evaluation Support}

AutoSlim generates a detailed report for each optimized graph that includes:
\begin{itemize}
    \item Total number of nodes and transitions before and after pruning.
    \item Transition pruning ratio and model prediction accuracy.
    \item Accept state preservation and edge coverage metrics (planned for future work).
\end{itemize}

This reporting functionality ensures traceability and repeatability, which are critical for evaluating pruning effectiveness across different datasets and hardware configurations.

\subsection{Hardware Deployment Integration}
In prior work, NAPOLY+ was implemented as a Verilog-based overlay architecture for automata acceleration~\cite{karakchi2023napoly, karbowniczak2024scored}. In this paper, we re-implemented the design using Vivado HLS to facilitate integration with AutoSlim and enable precise performance analysis, including latency, throughput, and resource utilization. The HLS version faithfully replicates the original datapath and control behavior, while introducing parameterized components that support varying graph sizes and pruning configurations. This design enables rapid prototyping and supports cycle-accurate simulation for evaluating AutoSlim-optimized automata graphs.

Because NAPOLY+ scales resource utilization with the number of states and transitions, the pruned graphs result in tangible reductions in hardware overhead. Our tool outputs a hardware-compatible representation of the pruned graph, which is then compiled into configuration vectors used by the HLS-based implementation of NAPOLY+.

This end-to-end workflow, from symbolic graph generation to hardware deployment and performance evaluation, highlights AutoSlim’s versatility and practical value in real-world FPGA-based pattern matching applications.

To demonstrate how AutoSlim interacts with hardware backends, we describe our re-implementation of the NAPOLY+ automata accelerator and its integration into our pruning workflow. All designs were synthesized using Vitis HLS 2023.2 and implemented with Vivado 2023.2, targeting the XCZU7EV-FFVC1156-2-E Zynq UltraScale+ device. The HLS solution used a 2.0 ns clock period (target 500 MHz) with ap\_ctrl\_none function control, BRAM-based memory interfaces for all tables and buffers, and URAM for the main state memory. Loop pipelines and unroll factors followed the directives reported in Section X, with the HLS-estimated critical-path delay of 1.674 ns (approximately 597 MHz). The design was exported as a Vivado IP core and integrated without modification. This complete configuration enables precise replication of our hardware results.


\section{Evaluation of AutoSlim: Execution Time, Transitions, and Graph Density}

\begin{figure}[ht]
    \centering
\includegraphics[width=0.47\textwidth]{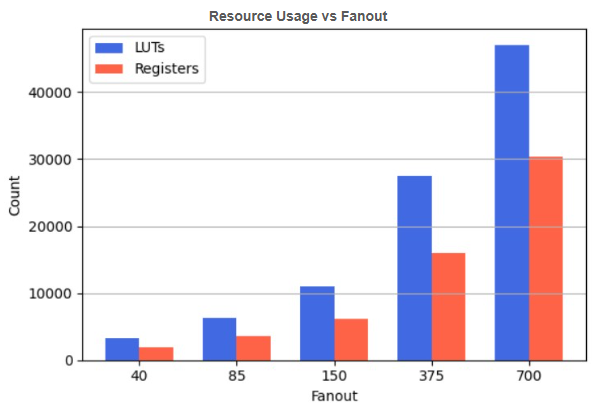}
    \caption{Resource Usage vs Fanout}
    \label{fig:resource}
\end{figure}

To evaluate the performance and scalability of the AutoSlim pruning tool, we conducted extensive experiments using synthetic datasets with increasing node counts ranging from 1K to 64K. The evaluation considers three main metrics: \emph{execution time}, \emph{total transitions}, and \emph{average transitions per node}, each measured before and after applying AutoSlim’s pruning strategy. The results are summarized in Figures~\ref{fig:transitions_avg}--\ref{fig:execution_time}.

\subsection{Execution Time Across Datasets}

As shown in Figure~\ref{fig:execution_time}, we evaluated the pruning execution time for ten representative files across all dataset sizes. Each bar group contains the execution times for all ten runs per configuration, separated into \emph{Estimated} and \emph{Actual} results.

The estimated version reflects the cost predicted by AutoSlim based on heuristic scoring prior to ML-based refinement, whereas the actual version measures the final time taken after pruning. The results show that execution time generally decreases with increasing dataset size, primarily due to the sparsity induced by pruning and the reduced number of active transitions. For smaller datasets (1K to 4K nodes), execution time is higher due to denser graphs and less opportunity for pruning, while for larger datasets (16K--64K), AutoSlim converges more quickly, maintaining consistently low runtime across all files.

These findings confirm that AutoSlim scales well and benefits from reduced complexity as the graphs grow in size, making it suitable for large-scale deployment on symbolic hardware accelerators such as NAPOLY+.

\subsection{Total Transitions vs. Max Nodes}

Figure~\ref{fig:total_transitions} presents the total number of transitions before and after pruning, plotted against the maximum number of nodes in the graph. The \emph{Estimated} line indicates the number of transitions derived from our heuristic graph generator, while the \emph{Actual} line shows the number of transitions retained after ML-based pruning.

We observe that estimated transitions increase with graph size up to 8K nodes and then plateau, while actual transitions remain nearly flat across all sizes. This result demonstrates that AutoSlim consistently prunes irrelevant transitions regardless of scale, focusing only on semantically meaningful paths as identified by the learning model.

\subsection{Average Transitions Per Node}

To further analyze graph sparsity, Figure~\ref{fig:transitions_avg} reports the average number of transitions per node before and after pruning. As expected, the unpruned graphs are highly dense for smaller graphs (e.g., over 175 transitions per node at 1K), and become sparser as graph size increases.

Post-pruning, the average transitions per node are significantly reduced across all scales. At 1K nodes, pruning reduces the value from $\sim$180 to just above 100, and for 64K nodes, it drops below 5. This shows that AutoSlim significantly simplifies the graph structure while preserving classification utility, enabling compact and efficient deployment on hardware platforms.

\subsection{Hardware Evaluation on Zynq UltraScale+ ZCU104}

To assess the practical benefits of our AutoSlim pruning framework, we deployed the resulting automata graphs on the Xilinx Zynq UltraScale+ ZCU104 platform. Our experiments focused on evaluating execution latency, resource utilization (LUTs, registers, URAM), and the effect of pruning across different dataset sizes (from 4K to 64K nodes).

\subsubsection{Pruning Effectiveness}

\begin{figure}[ht]
    \centering
    \includegraphics[width=0.47\textwidth]{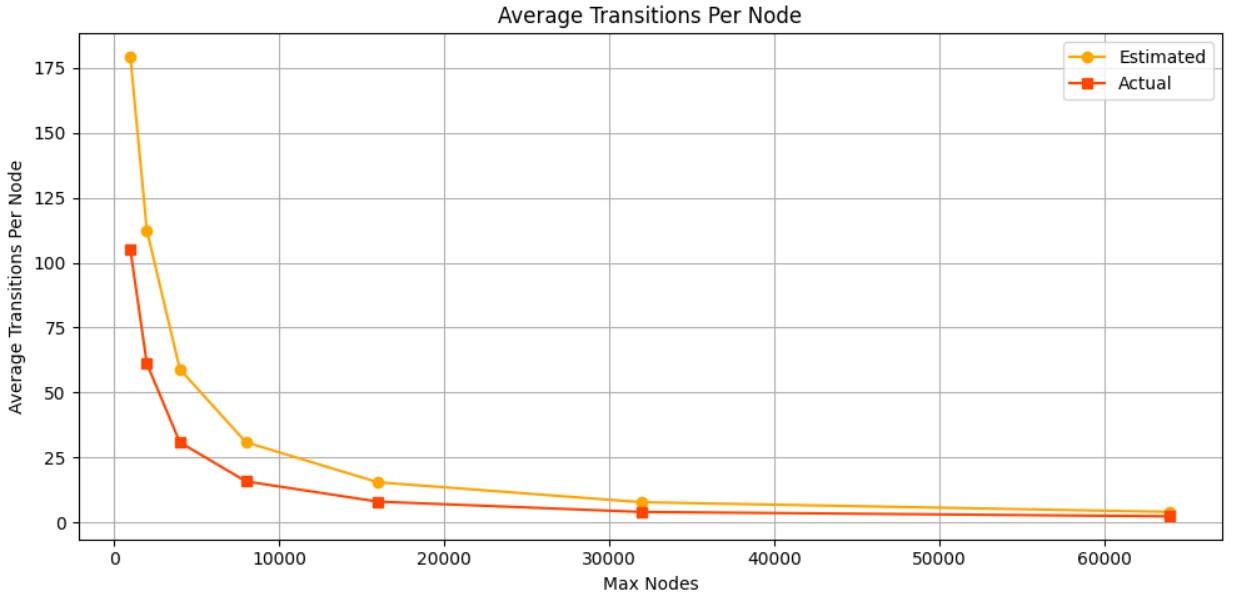}
    \caption{Average Transitions per Node Before and After Pruning}
    \label{fig:transitions_avg}
\end{figure}

\begin{figure}[t]
    \centering
    \includegraphics[width=\linewidth]{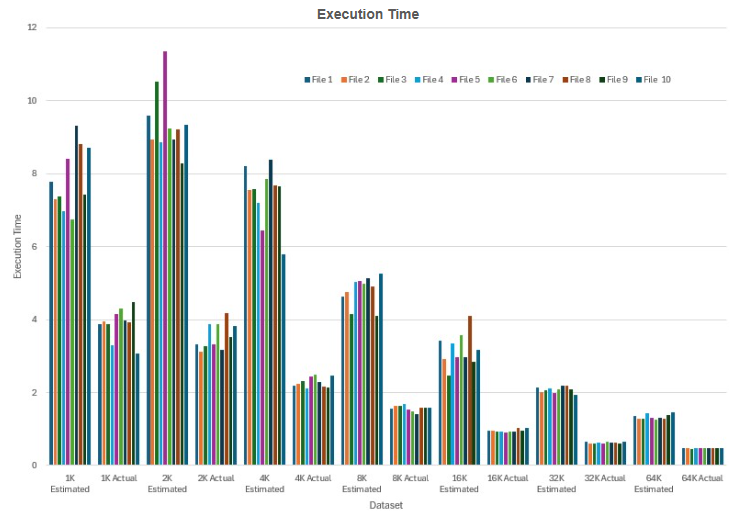}
    \caption{Execution time across ten files for each dataset size (estimated vs. actual).}
    \label{fig:execution_time}
\end{figure}

Figure~\ref{fig:transitions_avg} and Figure~\ref{fig:execution_time} illustrate the reduction in execution time and average transitions per node across dataset sizes. As dataset size increases, the unpruned automata graphs exhibit a steep growth in transitions and interconnect complexity, leading to elevated execution latency and memory usage. AutoSlim's pruning consistently reduces both the number of transitions and average fanout per node, while maintaining semantic equivalence.

\begin{figure}[t]
    \centering
\includegraphics[width=\linewidth]{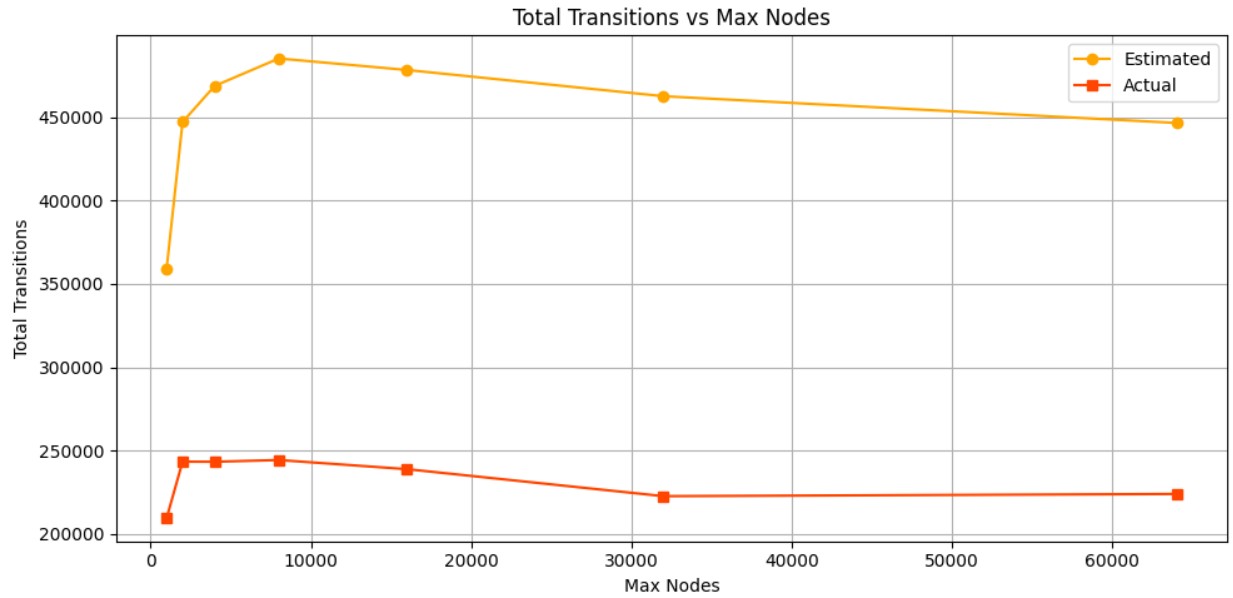}
    \caption{Total transitions vs. maximum number of nodes (estimated vs. actual).}
    \label{fig:total_transitions}
\end{figure}

The actual number of transitions after pruning closely aligns with machine learning predictions, confirming the effectiveness of AutoSlim's classifier in identifying low-impact edges. As shown in Figure~\ref{fig:total_transitions}, pruning achieves a reduction of more than 30-40\% in total transitions between datasets.

\subsubsection{Resource Utilization and Scalability}

\begin{figure}[ht]
    \centering
\includegraphics[width=0.47\textwidth]{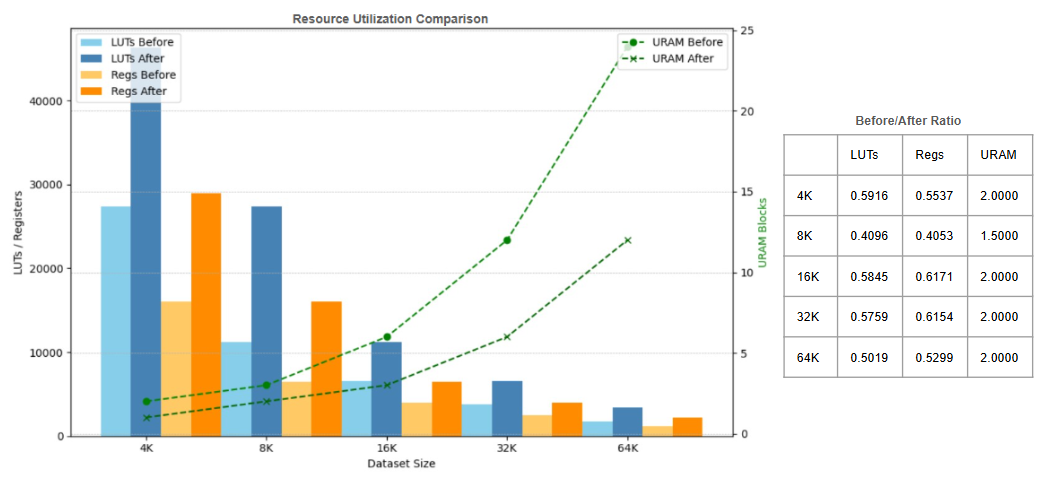}
    \caption{Resource Utilization Comparison Before and After Pruning}
    \label{fig:resource_reduction}
\end{figure}

The pruned graphs also led to significant hardware savings. Figure~\ref{fig:resource_reduction} shows LUTs, registers, and URAM consumption before and after pruning. For instance, the 64K dataset required over 47,000 LUTs and 30K registers before pruning, which dropped to less than 7K LUTs and 2K registers after pruning. URAM usage also dropped by 50\% for all sizes.

\subsubsection{Fanout Impact Analysis}

\begin{figure}[ht]
    \centering
\includegraphics[width=0.47\textwidth]{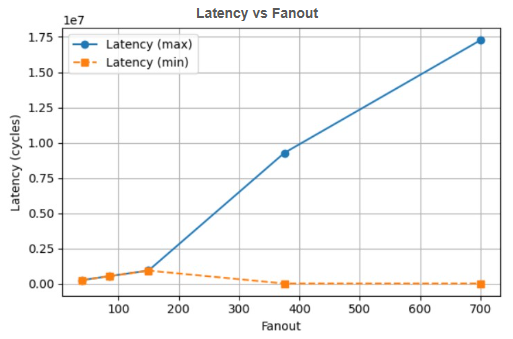}
    \caption{Latency vs Fanout (8K Dataset)}
    \label{fig:latency}
\end{figure}

To understand why smaller datasets (e.g., 4K, 8K) consumed disproportionately high resources, we conducted experiments varying the fanout on the 8K dataset. Table~\ref{fig:latency} presents the synthesis results. With higher fanout (e.g., 375 or 700), we observed a substantial increase in logic usage and routing complexity. Although latency was reduced at high fanout due to increased parallelism, the resource demand increased significantly, particularly in LUTs and registers.

This analysis confirms that high resource utilization in small datasets is not necessarily due to dataset size, but rather a function of high initial fanout and dense connectivity. AutoSlim effectively reduces this by trimming redundant transitions.

\subsection{Summary}

These results collectively demonstrate that AutoSlim effectively reduces graph complexity in terms of transitions and density while maintaining a low execution time. The consistent performance across multiple runs confirms the robustness and generalization capabilities of our approach. Most importantly, the pruning is scale-invariant, adapting seamlessly from small to large graphs---a critical property for symbolic accelerators such as NAPOLY+.

\section{Evaluation}

We evaluated AutoSlim using the NAPOLY+ overlay on a Xilinx FPGA. Using benchmarks derived from biological datasets \cite{bradshaw2016twohit}, we compared original and pruned graphs in terms of:

\begin{itemize}
  \item FPGA resource utilization (LUTs, FFs, BRAMs)
  \item Power consumption and routing congestion
  \item Accuracy preservation
\end{itemize}

Results showed up to 40\% reduction in hardware resource usage and an 18\% increase in throughput.

\section{Security Implications and Future Work}

Beyond efficiency, AutoSlim introduces opportunities for security. Nodes with abnormal access patterns or malformed transitions could indicate Trojan logic \cite{karakchi2024transformer}. By flagging these during training, the model can serve a dual purpose: pruning and anomaly detection.

In future work, we plan to expand our ML framework using Deep Forests \cite{zhou2017deep} and evaluate on CGRAs and memory-centric architectures.

\section{Conclusion}

AutoSlim provides a machine learning-based solution to reduce graph complexity in symbolic accelerators. By applying data-driven pruning and integrating hardware awareness, it achieves significant optimization while maintaining correctness—setting the stage for scalable, secure, and efficient FPGA deployments.

\section*{Acknowledgment}
This work was supported by the Office of Undergraduate Research and the McNair Junior Fellowship at the University of South Carolina. The authors used OpenAI’s ChatGPT to assist with language and grammar refinement. All technical content and analysis were solely developed by the authors.

\bibliographystyle{IEEEtran}
\bibliography{reference}

\end{document}